\documentclass[runningheads, orivec]{llncs}
\usepackage[T1]{fontenc}
\usepackage{graphicx}
\usepackage{color}
\usepackage{url}
\usepackage{xcolor}

\usepackage{amsmath}
\usepackage{hyperref}
\usepackage{amsfonts}
\usepackage{amssymb}
\usepackage{booktabs}
\usepackage{multirow}
\usepackage{makecell}
\usepackage[misc]{ifsym}
\begin{document}
\title{Full Integer Arithmetic Online Training for Spiking Neural Networks}
\author{Ismael Gomez \and Guangzhi Tang \Letter}
% Second Author\inst{2,3}\orcidID{1111-2222-3333-4444} 
\institute{Department of Advanced Computing Sciences, Maastricht University \\ Maastricht, The Netherlands\\
% , Princeton NJ 08544, USA \and
\email{guangzhi.tang@maastrichtuniversity.nl}}
\maketitle              % typeset the header of the contribution
\begin{abstract}
Spiking Neural Networks (SNNs) are promising for neuromorphic computing due to their biological plausibility and energy efficiency. However, training methods like Backpropagation Through Time (BPTT) and Real Time Recurrent Learning (RTRL) remain computationally intensive. This work introduces an integer-only, online training algorithm using a mixed-precision approach to improve efficiency and reduce memory usage by over 60\%. The method replaces floating-point operations with integer arithmetic to enable hardware-friendly implementation. It generalizes to Convolutional and Recurrent SNNs (CSNNs, RSNNs), showing versatility across architectures. Evaluations on MNIST and the Spiking Heidelberg Digits (SHD) dataset demonstrate that mixed-precision models achieve accuracy comparable to or better than full-precision baselines using 16-bit shadow and 8- or 12-bit inference weights. Despite some limitations in low-precision and deeper models, performance remains robust. In conclusion, the proposed integer-only online learning algorithm presents an effective solution for efficiently training SNNs, enabling deployment on resource-constrained neuromorphic hardware without sacrificing accuracy.

\end{abstract}

\section{Introduction}

Spiking Neural Networks (SNNs) have emerged as a promising class of energy-efficient models inspired by the dynamics of biological neurons. Their event-driven and sparse communication mechanism enables significantly reduced power consumption, particularly suited for deployment in neuromorphic hardware \cite{nguyen_review_2021}. This makes SNNs ideal for edge computing and real-time applications involving spatiotemporal data. Furthermore, SNN inference can be made even more efficient through quantization techniques, such as Post-Training Quantization (PTQ) and Quantization-Aware Training (QAT) \cite{krishnamoorthi_quantizing_2018,schnoll_fast_2023}. However, training SNNs remains computationally expensive. Standard methods like Backpropagation Through Time (BPTT) are memory-intensive and unsuitable for online and low-power learning scenarios \cite{wu_spatio-temporal_2018}.

To mitigate the limitations of BPTT, several online training algorithms have been proposed that approximate gradient-based updates in a forward-in-time approach \cite{xiao_online_2022,bellec_solution_2020}. These methods leverage local learning rules and eligibility traces to support online learning with constant memory consumption. However, these approaches still rely on high-precision floating-point arithmetic, which imposes substantial computational and memory overhead, especially when deployed on energy-constrained neuromorphic platforms. The high-precision floating-point format requires high memory and computational resources, limiting the efficiency of these algorithms in practice \cite{horowitz_11_2014}.

Integer-only learning provides an attractive alternative, offering drastic reductions in energy consumption and hardware complexity. Recent work has shown that training Artificial Neural Networks (ANNs) using integer arithmetic is feasible with careful rescaling, clipping, and architectural modifications \cite{ghaffari_is_2023,song_pocketnn_2022,pirillo_nitro-d_2024}. However, extending these ideas to SNNs is non-trivial because of how the temporal dynamics and spike-based computation influence the quantization noise and numerical instability introduced by integer-based training. Approaches like \cite{zou_all_2024} employ integer quantization in the ANN2SNN conversion. Nonetheless, the training process includes floating-point values in the encoding and optimization steps. In \cite{luo_integer-valued_2024}, integer-based values are used during training through their I-LIF neuron. However, the training process does not employ integer-only arithmetic, as integer operations are only applied to activations of the I-LIF neuron. As a result, SNN training with integer-only arithmetic remains underexplored.

In this paper, we propose an online training algorithm for SNNs that uses only integer arithmetic\footnote{Code: https://github.com/ERNIS-LAB/Integer-Arithmetic-SNN-Learning}. To solve the gradient precision challenge, we introduce two sets of weights: high-precision shadow weights for gradient updates and low-precision weights for inference and error propagation. The online training algorithm uses presynaptic and postsynaptic traces for local gradient approximation while constraining all forward and backward computations to low-precision integers. We evaluated the proposed method across three SNN architectures (fully connected, convolutional, and recurrent) on the MNIST and SHD datasets. Experimental results show that our approach achieves competitive accuracy compared to floating-point baselines while significantly reducing memory footprints and computational costs.

\section{Methodology}

\subsection{Mixed-Precision Integer Training}

One of the primary challenges of low-precision integer training is managing the scale of weight updates. Low-precision integer weights and gradients constrain the minimal update step, leading to coarse adjustments that can undermine the stability of the gradient descent process. To mitigate this issue in SNN online training, we introduce a mixed-precision approach that employs two distinct weight representations: $W_{Sh}$ \textbf{Shadow weights} (High-precision integers), store high-precision gradients during training; and, $W_{LP}$ \textbf{Low-precision weights} (Low-precision integers), used for forward and backward computations.

\begin{figure}[htbp]
\centering
\includegraphics[width=0.8\textwidth]{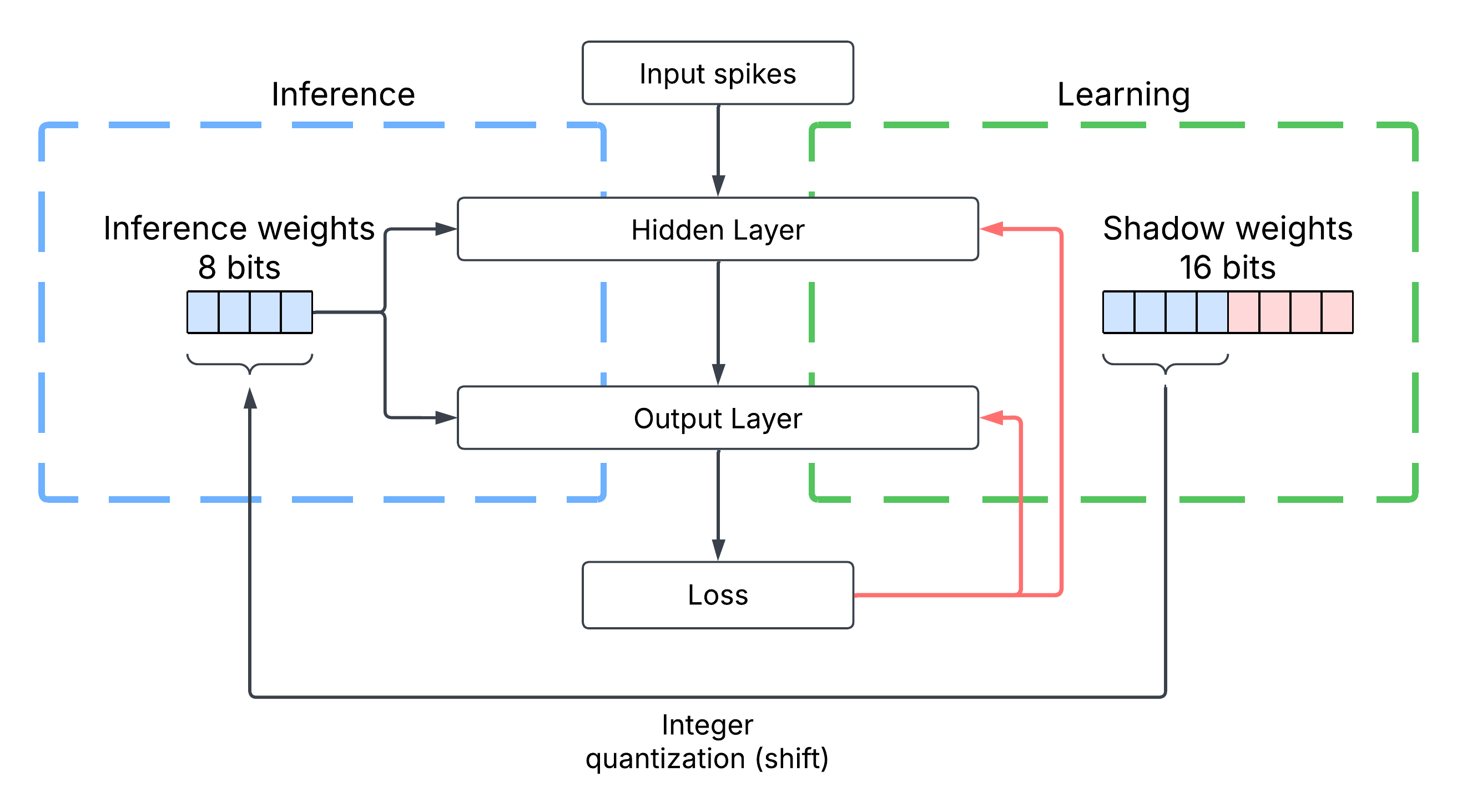}
\caption{Overview of the mixed-precision integer training process. Low-precision integer weights are used for the forward pass and the gradient calculation in the backward pass. The weights update is made on the high-precision shadow weights, which are then bit-shifted to regenerate the low-precision weights for the next iteration.}
\label{fig:integer_training}
\end{figure}

In each training iteration, the forward and backward propagations are computed using low-precision integer weights, reducing the computational cost of training. The computed gradients are used to update the shadow weights, where high precision allows for the preservation of update fidelity. Before starting a new training iteration, the low-precision weights are updated by applying bit-shifting to the shadow weights (Figure \ref{fig:integer_training}). The integer-mixed-precision approach (MP) balances precision and efficiency, enabling integer-only arithmetic while stabilizing gradient descent. Compared to the full-precision floating-point model (FP32), where 32-bit precision weights were used in every step, our MP approach reduces memory footprints and computational costs.

\subsection{Online SNN Learning Algorithm}
Our algorithm is gradient-based, local, online, and uses integer-only arithmetic. Built on top of existing gradient-based online learning method for SNN \cite{tang_biograd_2021}, it uses local eligibility traces to approximate the gradients required for learning in SNN and update the weights. Our full integer arithmetic online training supports fully connected, convolutional, and recurrent SNNs.

\subsubsection{Forward Step}

We use the leaky-integrate-and-fire (LIF) neurons in the SNN. The voltage decay of LIF is converted to a right bit-shift operation $\gg$ adjusted with the voltage decay rate $\beta$ to maintain the integer-only suitability. At step $t$, the neuron's voltage $V$ from layer $i$ is updated based on the input spikes $s$ and forward low-precision weight $\mathbf{W}_{LP}^{(i)}$, as shown in Equation \ref{eq:lif_update_int}.
\begin{equation}
    \hat{\beta} = \lfloor \log_2(1/\beta) \rfloor 
    \label{eq:bit_shift}
\end{equation}
\begin{equation}
    \mathbf{V}^{(i)}(t) = \mathbf{V}^{(i)}(t-1) \gg \hat{\beta} + \mathbf{W}_{LP}^{(i)} \mathbf{s}^{(i)}(t)
    \label{eq:lif_update_int}
\end{equation}
A surrogate gradient $\widetilde{\nabla V}$ is computed in Equation \ref{eq:pseudo_grad} by assigning a positive value to those voltages whose values are close to the firing threshold $V_{th}$.
\begin{equation}
    \widetilde{\nabla V}^{(i)}(t) = | V^{(i)}(t) - V_{th}^{(i)} | < Grad_{win}^{(i)}
    \label{eq:pseudo_grad}
\end{equation}
Our method utilizes two traces to capture information about how the weights influence the neuron's output: A presynaptic spike trace $T_{pre}$, which reflects the history of presynaptic neuron spiking (Equation \ref{eq:trace_pre}), and a correlation trace $T_{corr}$, which captures the correlation between pre- and postsynaptic neuron activities (Equation \ref{eq:trace_weight}). The idea of eligibility traces was first introduced by RFLO \cite{murray2019local} and adapted by E-prop for SNNs \cite{bellec_solution_2020}. This work introduces full integer arithmetic computation for eligibility-trace-based online SNN learning, where the correlation trace is similar to the eligibility trace in E-prop. Details of the two traces can be found in \cite{tang_biograd_2021}.
\begin{equation}
    T_{pre}^{(i)}(t) = T_{pre}^{(i)}(t-1) \gg \hat{\beta} + \mathbf{s}^{(i)}(t)
    \label{eq:trace_pre}
\end{equation}
\begin{equation}
    T_{corr}^{(i)}(t) =  T_{corr}^{(i)}(t-1) + T_{pre}^{(i)}(t) \cdot \widetilde{\nabla V}^{(i)}(t)
    \label{eq:trace_weight}
\end{equation}

\subsubsection{Loss Calculation}

During the inference, the output spikes from each time step are aggregated, and the prediction of the network is computed by choosing the neuron with the maximum number of spikes at the end (one-hot encoded). The loss function simplifies the softmax function to work with integer-only arithmetic by removing the exponential operations and introducing a bit-shifting by approximating the number of computation steps $t_s$ to the closest power of 2. The one-hot encoded true label $y$ is multiplied by the loss precision $\alpha$ to regulate the range of values.
\begin{equation}
    error^{pred} = (prediction \cdot \alpha) \, \gg \, \lfloor \log_2(t_s) \rfloor \, - y \cdot \alpha
    \label{eq:loss}
\end{equation}

\subsubsection{Error Propagation}

In the hidden layers, the layer $i$ error voltage $V^{(i)}_{fb}$ is obtained by propagating the prediction error using the feedback weights $W^{(i)}_{fb}$, which correspond to the transposed low-precision weights of layer $i+1$. This method simplifies error propagation, reducing memory usage and computational demands while maintaining efficient error propagation. In the output layer, this error corresponds to the prediction error.
\begin{equation}
    V^{(i)}_{fb} = W_{fb}^{(i)} \times error^{pred}
    \label{eq:direct_hidden}    
\end{equation}

\subsubsection{Weights Update}

The layer feedback voltage is multiplied by the correlation trace, indicating which neurons had more relevance in the prediction to change their value. This product is then summed across the batch dimension, producing the low-precision gradient $\Delta^{(i)}$ (Equation \ref{eq:delta_hidden}). Next, the shadow weights are updated by subtracting the scaled gradient, where the layer-specific learning rate $\eta^{(i)}$ is applied via right bit-shifting (Equation \ref{eq:weight_update}). A weight decay term $\rho^{(i)}$, also applied via bit-shifting, is subtracted to regularize the weights. These updates are applied to the high-precision shadow weights, which are then quantized to generate the low-precision weights used in the next training iteration.
\begin{equation}
    \Delta^{(i)}(t) = \sum_{b=1}^{B} [V^{(i)}_{fb} \cdot T_{corr}^{(i)}(t)]
    \label{eq:delta_hidden}
\end{equation}
\begin{equation}
    W_{Sh}^{(i)} =  W_{Sh}^{(i)} - \Delta^{(i)}(t) \gg \hat{\eta^{(i)}} - W_{Sh}^{(i)} \gg \hat{\rho^{(i)}}
    \label{eq:weight_update}    
\end{equation}

\subsubsection{Initialization of the weights}

The integer weight initialization needs to ensure numerical stability and preserve the relative scale of the weights between layers. We first perform a standard floating-point weight initialization. The maximum absolute value across all weights is identified to preserve the relative scale of weights after quantization. Each weight matrix is then quantized proportionally to this global maximum. The quantization process follows a uniform scheme, where weights are scaled by a step size determined by the global maximum and the bit-width constraint.

\subsubsection{Gradient clipping}
Training stability is one of the main challenges in integer-only training \cite{pirillo_nitro-d_2024,zhao_distribution_2021} and gradient clipping has been widely used in deep learning to obtain smoother optimization \cite{zhao_distribution_2021}. We apply gradient clipping with a global clipping parameter $\Delta_{\text{max}}$, assigned after hyperparameter search, for all the layers.
\begin{equation}
\Delta = \max(-\Delta_{\text{max}}, \min(\Delta, \Delta_{\text{max}}))
    \label{eq:gradient_clip}
\end{equation}

\subsection{Convolutional Spiking Neural Networks}

\begin{figure}[htbp]
\centering
\includegraphics[width=0.8\textwidth]{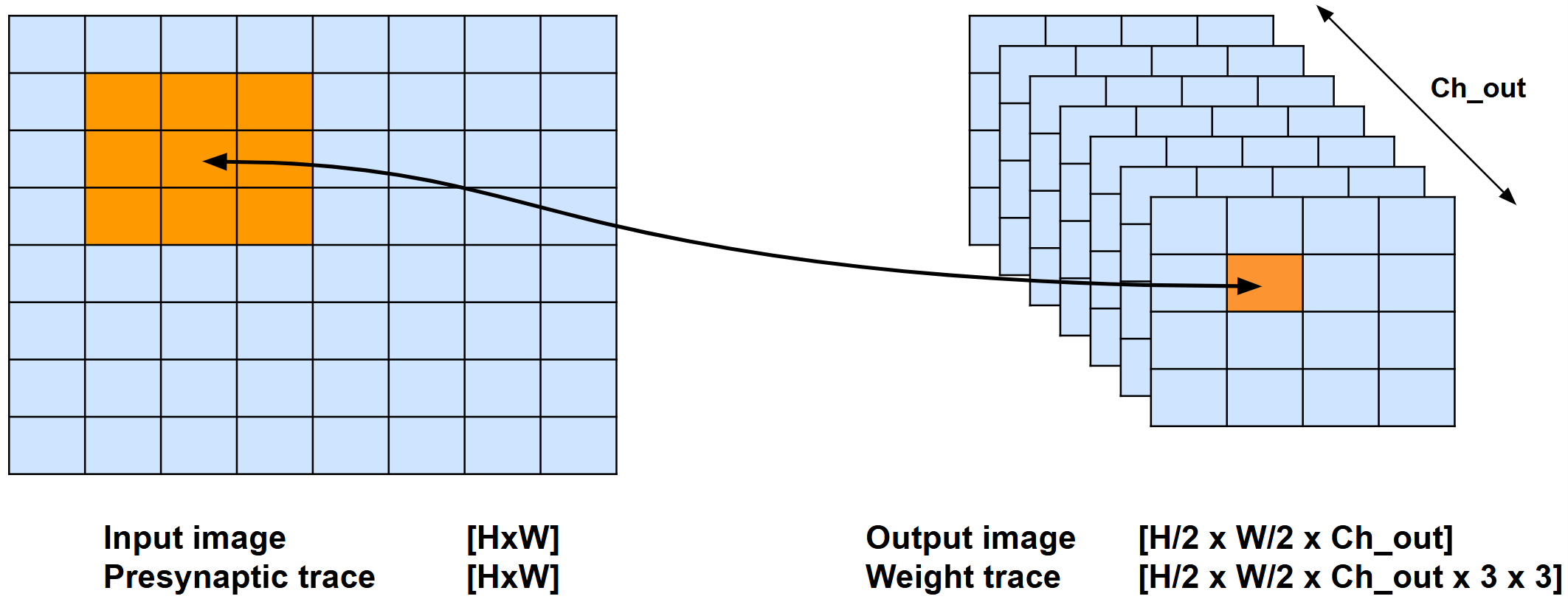}
\caption{Diagram of how the correlation trace $T_{corr}$ is calculated in a CSNN.}
\label{fig:cnn_trace}
\end{figure}

We adapted our algorithm for Convolutional SNNs (CSNN). This involves changing the computation of the correlation trace due to the weight sharing of convolutional operations. As shown in Figure \ref{fig:cnn_trace}, $T_{corr}$ is computed by taking, for each output value, the kernel values used in each convolution ($unfold$ operation). Those values are then multiplied by the surrogate gradient (Equation \ref{eq:trace_weight_cnn}).
\begin{equation}
    T_{corr}^{(i)}(t) =  T_{corr}^{(i)}(t-1) + \operatorname{unfold}(T_{pre}^{(i)}(t)) \cdot \widetilde{\nabla V}^{(i)}(t)
    \label{eq:trace_weight_cnn}
\end{equation}
The other major change in the CSNN implementation is in the weight update step. The feedback weights used in the SNN case are substituted by a transposed convolutional operation that transforms the prediction error (dimension B $\times$ $Output_{dim}$) to be multiplied properly with the correlation trace (dimension B $\times$ $Ch_{out}$ $\times$ H/2 $\times$ W/2 $\times$ $Ch_{in}$ $\times$ kernel\_size $\times$ kernel\_size) (Equation \ref{eq:error_prop_csnn}).
\begin{equation}
    v^{(i)}_{fb} = Conv^{T (i)} (error^{pred})
    \label{eq:error_prop_csnn}
\end{equation}

\subsection{Recurrent Spiking Neural Networks}

Implementing the recurrent connection in the spiking neural network implies a change in how the voltage is updated. Each layer has an extra set of weights $\mathbf{W}_{rec}$ that process the voltage from the previous time step. To prevent the voltage from uncontrolled growth, a bit-shifting operation $\mathcal{B}_{16}$ is applied to the voltage and only the first 16 bits of information are taken.
\begin{equation}
    \mathbf{V}^{(i)}(t) = \beta \cdot \mathbf{V}_{\text{s}}^{(i)}(t-1) + \mathbf{W}^{(i)} \mathbf{s}^{(i)}(t) + \mathbf{W}^{(i)}_{rec}\, \mathcal{B}_{16}\big(\mathbf{V}^{(i)}(t-1)\big)
    \label{eq:lif_update_rec}
\end{equation}
Instead of calculating specific traces and updating the recurrent weights, a random recurrent feedback approach has been followed in this work. Random recurrent mechanisms replace the need for training recurrent weights by using fixed, random matrices to process the previous voltage. The effectiveness of random feedback techniques has been demonstrated in different tasks inside the training of deep neural networks \cite{lillicrap_random_2014,fernandez_gradient-free_2024}.

\section{Experiments and Results}
\subsection{Experimental Setup}
All experiments were conducted by training the models for 50 epochs. A training batch size of 128 was used in all the experiments. To mitigate the effects of stochasticity in training, all reported test accuracies represent the mean and standard deviation computed over 10 runs with different random seeds.

The learning algorithm was first tested on the MNIST dataset \cite{lecun_gradient-based_1998}. To use the temporal dynamics of the SNNs in this static dataset, the images fed to the network during the forward pass were the result of a spike-rate encoding, the same as the encoding method defined in \cite{tang_biograd_2021}. The spike time and number of hidden neurons were set to 10 and 100, respectively.

The Spiking Heidelberg Digits (SHD) dataset \cite{cramer_heidelberg_2022} is a benchmark for temporal neuromorphic tasks, consisting of spiking data generated from audio recordings of spoken digits pronounced by 12 speakers. There are 20 labels corresponding to the 10 digits in English and German. Given the variability in recording lengths, ranging from 0.2 to 1.4 seconds, the spike events are aggregated into 10 uniform time frames per sample. To address the sparsity of the data and increase efficiency, the 700 channels are further grouped into sets of 4, reducing the effective channel count to 175 while preserving essential temporal and spectral information. The spike time and number of hidden neurons were set to 10 and 256, respectively.

\subsection{Fully Connected and Convolutional SNN on MNIST}

\paragraph{Fully Connected Spiking Neural Network}
Table \ref{tab:mnist_snn} shows the results for different MP configurations of a single-layer SNN. Surprisingly, the MP-16-8 model outperforms the FP32 baseline. Gradient clipping and weight decay techniques applied in the MP configurations were not applied in the FP32 model, potentially giving MP-16-8 an advantage. Additionally, quantization can act as implicit regularization, helping generalization by reducing overfitting \cite{banner_scalable_2018}. This might also explain why MP-16-16 underperforms slightly despite higher inference precision.

\begin{table}[h]
    \centering
    \caption{Performance comparison on MNIST between different mixed-precision configurations. FP32 represents the floating-point full precision configuration.}
    \begin{tabular}{l|ccc|c}
        \toprule
         & \multicolumn{3}{c|}{\textbf{Shadow weights}} & \multirow{2}{*}{\textbf{FP32}} \\
        \cmidrule(lr){2-4}
        \makecell{\textbf{Inference} \\ \textbf{weights}} & \textbf{4 bits} & \textbf{8 bits} & \textbf{16 bits} &  \\
        \midrule
        \textbf{4 bits}  & 65.97 $\pm$ 0.34 \% & 94.24 $\pm$ 0.27 \% & 95.47 $\pm$ 0.15 \% & \multirow{4}{*}{97.33 $\pm$ 0.15 \% } \\
        \textbf{8 bits}  & - & 96.89 $\pm$ 0.14 \% & 97.55 $\pm$ 0.09 \% &  \\
        \textbf{16 bits} & - & - & 97.43 $\pm$ 0.14 \% &  \\
        \textbf{FP Biograd \cite{tang_biograd_2021}} & - & 96.72 $\pm$ 0.20 \% & 98.10 $\pm$ 0.07 \% & \\
        \bottomrule
    \end{tabular}
    \label{tab:mnist_snn}
\end{table}

The MP-8-4 model shows the low accuracy (94.24\%), highlighting the information loss with 4-bit inference weights. Increasing shadow precision to 16 bits improves performance (95.47\%), confirming its importance for gradient information storage.

Notably, using 4-bit shadow weights (MP-4-4) results in a substantial drop in accuracy, significantly lower than that of all other configurations. This performance degradation underscores the critical role of maintaining high shadow weight precision during the weight update phase.

Additionally, we compared the results with BioGrad~\cite{tang_biograd_2021}, an online SNN training method using low-precision weights for inference and with high-precision floating point arithmetic for backward updates during training. With 8-bit and 16-bit inference, BioGrad achieves results comparable to our MP configurations. This highlights the efficiency of our integer-only approach.

\paragraph{Convolutional Spiking Neural Network}
Table \ref{tab:mnist_cnn} reports results for the CSNN model on MNIST, using a kernel size of $5 \times 5$, 32 filters, stride 2, and no padding.

\begin{table}[ht]
    \centering
    \caption{Performance comparison on MNIST between different mixed-precision configurations. FP32 represents the floating-point full precision configuration.}
    \begin{tabular}{l|cc|c}
        \toprule
         & \multicolumn{2}{c|}{\textbf{Shadow weights}} & \multirow{2}{*}{\textbf{FP32}} \\
        \cmidrule(lr){2-3}
        \textbf{Inference weights} & \textbf{8 bits} & \textbf{16 bits} &  \\
        \midrule
        \textbf{4 bits}  & 95.19 $\pm$ 0.84 \% & 97.73 $\pm$ 0.19 \% & \multirow{3}{*}{97.85 $\pm$ 0.12 \%} \\
        \textbf{8 bits}  & 97.16 $\pm$ 0.20 \% & 98.10 $\pm$ 0.10 \% &  \\
        \textbf{16 bits} & - & 97.88 $\pm$ 0.17 \% &  \\
        \bottomrule
    \end{tabular}
    \label{tab:mnist_cnn}
\end{table}

For convolutional layer training, MP-16-8 slightly outperforms FP32, suggesting implicit regularization in the use of low-precision weights. MP-8-4 shows a noticeable drop, confirming the limitations of low-precision inference.

CSNNs show better resilience to quantization than SNNs. The smaller performance drop in MP-16-4 and MP-8-8 suggests that convolutional structures benefit from weight sharing and local receptive fields, which help counteract the effects of quantization noise.

\subsection{Fully Connected and Recurrent SNN on SHD}

The results are compared with the \cite{cramer_heidelberg_2022}, which obtained the highest performances in the state-of-the-art with classic SNNs and RSNNs. The SNN and RSNN models in that work are trained using surrogate gradient descent in combination with BPTT. Table \ref{tab:results_shd} reports the performance of the MP configurations and a comparison with the FP32 configuration.

\begin{table}[ht]
    \centering
    \scriptsize
    \caption{Comparison of accuracy on SHD between the Mixed-Precision Integer and the FP32 approaches for the different models.}
    \label{tab:results_shd}
    \begin{tabular}{l|cccc|c}
        \toprule
        \textbf{Model} & \textbf{MP 16-4} & \textbf{MP 16-8} & \textbf{MP 16-12} & \textbf{MP 16-16} & \textbf{FP32} \\
        \midrule\midrule
        SNN & 49.85 $\pm$ 1.29 \% & 50.10 $\pm$ 1.14 \%  & 62.06 $\pm$ 1.16 \% & 61.92 $\pm$ 1.53 \% & 55.27 $\pm$ 1.97 \%\\
        SNN \cite{cramer_heidelberg_2022} & - & -  & - & - & 48.10 ± 1.60 \% \\
        RSNN & 57.62 $\pm$ 0.95 \% & 64.63\ $\pm$ 1.49 \% & 70.50 $\pm$ 1.43 \% & 67.75 $\pm$ 1.34 \% & 71.64 ± 0.95 \%\\
        RSNN \cite{cramer_heidelberg_2022} & - & - & - & - & 71.40 ± 1.90 \%\\
        \bottomrule
    \end{tabular}
\end{table}

\paragraph{Fully Connected Spiking Neural Network}
For the SNN model, the MP-16-12 (62.06\%) and MP-16-16 (61.92\%) configurations outperform FP32 (55.27\%), which could be attributed to quantization advantages on regularization similar to those observed in the MNIST experiments. The lack of gradient clipping and weight decay in the FP32 model, as well as the impact of fine-tuning, may have had a greater impact in this dataset. Notably, all mixed-precision configurations exceed the accuracy of the BPTT-trained baseline (48.1\%), demonstrating the efficacy of the proposed online learning approach in SNNs.

At lower inference precisions, performance drops significantly, with MP-16-4 (49.85\%) and MP-16-8 (50.1\%) matching the baseline but struggling to maintain the competitive accuracy of the higher-precision configurations. This highlights the complexity of the SHD dataset, where precise spike-timing information is crucial for classification. Unlike MNIST, which primarily relies on spatial feature encoding, SHD depends heavily on temporal feature representations, making it more sensitive to quantization-induced distortions.

\paragraph{Recurrent Spiking Neural Network}

In the RSNN model, a similar trend is observed, but with some distinctions due to recurrent connections. The MP-16-12 (70.50\%) and MP-16-16 (67.75\%) configurations closely matched the FP32 configuration (71.64\%) and the BPTT-trained baseline (71.40\%), confirming that random recurrent connections can achieve near-optimal performance while reducing computational training costs.

At lower precisions, the RSNN model experiences a significant accuracy drop (57.6\% in MP-16-4, 64.63\% in MP-16-8). The performance decline at lower bit widths is amplified in RSNNs, due to the recurrent connections. As they introduce additional weights and information into the voltage, the lower precision of the elements results in a bigger loss of information. These findings emphasize that recurrent architectures require greater precision in inference weights than feedforward SNNs to maintain effective temporal modeling.

\subsection{Efficiency of the Algorithm}

\paragraph{Memory usage} The improvements in memory load in the mixed-precision configuration are given by the reduced cost of storing lower-precision numbers. Additionally, assuming the algorithm is run in integer-based hardware, the components can be stored in flexible precisions (e.g., 9, 14 bits) instead of being constrained to standard 8, 16, or 32-bit formats. In Table \ref{tab:memory_cost}, the memory footprint of the MP-16-8 and FP32 SNN models trained on SHD is compared. The MP-16-8 model utilizes only 36.95\% of the memory used by the FP32 model.

\paragraph{Computational cost} The computational cost of the algorithm is determined by the total number of operations required for forward, backward propagation, and weight updates. Table \ref{tab:computational_cost} shows the total number of operations in the MP-16-8 and FP32 SNN models trained on SHD. Mixed-precision integer arithmetic significantly reduces this cost compared to traditional floating-point computations. Based on hardware measurements on a digital circuit with 45nm technology node \cite{horowitz_11_2014}, an 8-bit integer multiplication (0.2 pJ) is 18x more energy-efficient than a 32-bit floating-point multiplication (3.7 pJ). Besides, while an 8-bit integer addition (0.03 pJ) is 30x more efficient than its floating-point counterpart (0.9 pJ).

\begin{table}[ht]
    \centering
    \caption{Total memory usage of the algorithm in bytes. Calculated for the SNN model trained on the SHD dataset. Static memory refers to memory fixed at compile-time, it has faster access and is not modified during execution. Dynamic memory allocates elements that are not needed in every stage of the process and do not have to be saved at the end of execution. Applied to this algorithm, the static memory consists of the network weights and the layer and training parameters. Although the weights are updated at the end of every training iteration, they are constant during the inference and update steps. Dynamic memory comprehends the elements that are updated during the training iteration and their final value is not needed for the next iteration.}
    \begin{tabular}{l@{\hspace{15pt}}c@{\hspace{15pt}}c}
        \toprule
        \textbf{Model Section} & \textbf{MP 16-8} & \textbf{FP32} \\
        \midrule\midrule
        Static memory & 149,775 & 399,396\\
        Dynamic memory & 18,241,408 & 49,372,992\\
        \midrule
        \textbf{Total memory} & \textbf{18,391,383} & \textbf{49,772,388}\\
        \bottomrule
    \end{tabular}
    \label{tab:memory_cost}
\end{table}

\begin{table}[ht]
    \centering
    \caption{Total computational cost (number of operations) of the algorithm. Calculated for the SNN model and the SHD dataset. ADD - Addition, MUL - Multiplication, B-MUL - Binary Multiplication, EXP - Exponentiation.}
    \begin{tabular}{l@{\hspace{12pt}}c@{\hspace{12pt}}c@{\hspace{12pt}}c@{\hspace{12pt}}c@{\hspace{12pt}}c}
        \toprule
        \textbf{Model Configuration} & \textbf{ADD} & \textbf{MUL} & \textbf{B-MUL} & \textbf{Bit-shift} & \textbf{EXP} \\
        \midrule\midrule
        Floating-point & 35,240,315 & 34,315,325 & 706,560 & - & 617\\
        Integer-only & 34,911,751 & 33,553,745 & 706,560 & 1,245,173 & -\\
        \bottomrule
    \end{tabular}
    \label{tab:computational_cost}
\end{table}

Additional energy savings are achieved by the use of bit-shifting operations. In the integer-only version, tasks such as applying the voltage decay, learning rate, or weight decay rate are implemented using bit shifts instead of multiplications. Bit-shifting is virtually cost-free in hardware, requiring only simple register shifts instead of full multiplier circuits \cite{you_shiftaddnet_2020}. By leveraging both mixed-precision and integer arithmetic, the proposed method substantially reduces the energy required in the training and inference of SNNs.

\section{Conclusion and Discussion}

In this paper, we introduce an integer-only online learning approach for training SNNs, relying exclusively on fundamental integer arithmetic operations such as addition, multiplication, and bit shifting. Our method achieves competitive performance compared to floating-point counterparts. We validate the approach across various SNN architectures, highlighting the potential of a fully integer-based, local, and online learning framework for SNNs.

A key innovation of our method is its compatibility with integer-only arithmetic, intentionally limiting arithmetic operations to enhance learning efficiency. By combining the benefits of mixed-precision representation with integer-only arithmetic, the proposed algorithm significantly improves memory efficiency and reduces computational cost, enabling hardware-friendly implementation. Given that many digital neuromorphic processors rely on integer arithmetic \cite{davies_loihi_2018} or support accelerated integer operations \cite{gonzalez2024spinnaker2}, our approach is particularly suited for efficient SNN training on such platforms.

\subsubsection{\ackname} This publication is part of the project Brain-inspired MatMul-free Deep Learning for Sustainable AI on Neuromorphic Processor with file number NGF.1609.243.044 of the research programme AiNed XS Europe which is (partly) financed by the Dutch Research Council (NWO) under the grant https://doi.org/10.61686/MYMVX53467.

\bibliographystyle{splncs04}
\bibliography{references}

\begin{thebibliography}{10}
\providecommand{\url}[1]{\texttt{#1}}
\providecommand{\urlprefix}{URL }
\providecommand{\doi}[1]{https://doi.org/#1}

\bibitem{nguyen_review_2021}
Nguyen, D.A., Tran, X.T., Iacopi, F.: A {Review} of {Algorithms} and {Hardware} {Implementations} for {Spiking} {Neural} {Networks}. Journal of Low Power Electronics and Applications  \textbf{11}(2), ~23 (Jun 2021). \doi{10.3390/jlpea11020023}, \url{https://www.mdpi.com/2079-9268/11/2/23}, number: 2 Publisher: Multidisciplinary Digital Publishing Institute

\bibitem{krishnamoorthi_quantizing_2018}
Krishnamoorthi, R.: Quantizing deep convolutional networks for efficient inference: {A} whitepaper (Jun 2018). \doi{10.48550/arXiv.1806.08342}, \url{http://arxiv.org/abs/1806.08342}, arXiv:1806.08342

\bibitem{schnoll_fast_2023}
Schnöll, D., Wess, M., Bittner, M., Götzinger, M., Jantsch, A.: Fast, {Quantization} {Aware} {DNN} {Training} for {Efficient} {HW} {Implementation}. 2023 26th Euromicro Conference on Digital System Design (DSD) pp. 700--707 (Sep 2023). \doi{10.1109/DSD60849.2023.00100}, \url{https://ieeexplore.ieee.org/document/10456774/}, conference Name: 2023 26th Euromicro Conference on Digital System Design (DSD) ISBN: 9798350344196 Place: Golem, Albania Publisher: IEEE

\bibitem{wu_spatio-temporal_2018}
Wu, Y., Deng, L., Li, G., Zhu, J., Shi, L.: Spatio-{Temporal} {Backpropagation} for {Training} {High}-{Performance} {Spiking} {Neural} {Networks}. Frontiers in Neuroscience  \textbf{12}, ~331 (May 2018). \doi{10.3389/fnins.2018.00331}, \url{https://www.frontiersin.org/article/10.3389/fnins.2018.00331/full}

\bibitem{xiao_online_2022}
Xiao, M., Meng, Q., Zhang, Z., He, D., Lin, Z.: Online {Training} {Through} {Time} for {Spiking} {Neural} {Networks} (Dec 2022), \url{http://arxiv.org/abs/2210.04195}, arXiv:2210.04195

\bibitem{bellec_solution_2020}
Bellec, G., Scherr, F., Subramoney, A., Hajek, E., Salaj, D., Legenstein, R., Maass, W.: A solution to the learning dilemma for recurrent networks of spiking neurons. Nature Communications  \textbf{11}(1), ~3625 (Jul 2020). \doi{10.1038/s41467-020-17236-y}, \url{https://www.nature.com/articles/s41467-020-17236-y}

\bibitem{horowitz_11_2014}
Horowitz, M.: 1.1 {Computing}'s energy problem (and what we can do about it). In: 2014 {IEEE} {International} {Solid}-{State} {Circuits} {Conference} {Digest} of {Technical} {Papers} ({ISSCC}). pp. 10--14 (Feb 2014). \doi{10.1109/ISSCC.2014.6757323}, \url{https://ieeexplore.ieee.org/document/6757323}, iSSN: 2376-8606

\bibitem{ghaffari_is_2023}
Ghaffari, A., Tahaei, M.S., Tayaranian, M., Asgharian, M., Nia, V.P.: Is {Integer} {Arithmetic} {Enough} for {Deep} {Learning} {Training}? (Jan 2023). \doi{10.48550/arXiv.2207.08822}, \url{http://arxiv.org/abs/2207.08822}, arXiv:2207.08822

\bibitem{song_pocketnn_2022}
Song, J.S., Lin, F.: {PocketNN}: {Integer}-only {Training} and {Inference} of {Neural} {Networks} via {Direct} {Feedback} {Alignment} and {Pocket} {Activations} in {Pure} {C}++. ArXiv  (Jan 2022), \url{https://www.semanticscholar.org/paper/PocketNN%3A-Integer-only-Training-and-Inference-of-in-Song-Lin/81fcadee5f9fc3b57a35df38d4ce3eb9e868b426}

\bibitem{pirillo_nitro-d_2024}
Pirillo, A., Colombo, L., Roveri, M.: Nitro-d: Native integer-only training of deep convolutional neural networks. arXiv preprint arXiv:2407.11698  (2024)

\bibitem{zou_all_2024}
Zou, C., Cui, X., Feng, S., Chen, G., Zhong, Y., Dai, Z., Wang, Y.: An all integer-based spiking neural network with dynamic threshold adaptation. Frontiers in Neuroscience  \textbf{18} (Dec 2024). \doi{10.3389/fnins.2024.1449020}, \url{https://www.frontiersin.org/journals/neuroscience/articles/10.3389/fnins.2024.1449020/full}, publisher: Frontiers

\bibitem{luo_integer-valued_2024}
Luo, X., Yao, M., Chou, Y., Xu, B., Li, G.: Integer-{Valued} {Training} and {Spike}-{Driven} {Inference} {Spiking} {Neural} {Network} for {High}-performance and {Energy}-efficient {Object} {Detection} (Aug 2024). \doi{10.48550/arXiv.2407.20708}, \url{http://arxiv.org/abs/2407.20708}, arXiv:2407.20708

\bibitem{tang_biograd_2021}
Tang, G., Kumar, N., Polykretis, I., Michmizos, K.P.: {BioGrad}: {Biologically} {Plausible} {Gradient}-{Based} {Learning} for {Spiking} {Neural} {Networks} (Oct 2021), \url{http://arxiv.org/abs/2110.14092}, arXiv:2110.14092

\bibitem{murray2019local}
Murray, J.M.: Local online learning in recurrent networks with random feedback. Elife  \textbf{8},  e43299 (2019)

\bibitem{zhao_distribution_2021}
Zhao, K., Huang, S., Pan, P., Li, Y., Zhang, Y., Gu, Z., Xu, Y.: Distribution {Adaptive} {INT8} {Quantization} for {Training} {CNNs} (Feb 2021). \doi{10.48550/arXiv.2102.04782}, \url{http://arxiv.org/abs/2102.04782}, arXiv:2102.04782

\bibitem{lillicrap_random_2014}
Lillicrap, T.P., Cownden, D., Tweed, D.B., Akerman, C.J.: Random feedback weights support learning in deep neural networks (Nov 2014). \doi{10.48550/arXiv.1411.0247}, \url{http://arxiv.org/abs/1411.0247}, arXiv:1411.0247 [q-bio]

\bibitem{fernandez_gradient-free_2024}
Fernandez, J.G., Keemink, S., Gerven, M.v.: Gradient-{Free} {Training} of {Recurrent} {Neural} {Networks} using {Random} {Perturbations}. Frontiers in Neuroscience  \textbf{18},  1439155 (Jul 2024). \doi{10.3389/fnins.2024.1439155}, \url{http://arxiv.org/abs/2405.08967}, arXiv:2405.08967

\bibitem{lecun_gradient-based_1998}
Lecun, Y., Bottou, L., Bengio, Y., Haffner, P.: Gradient-based learning applied to document recognition. Proceedings of the IEEE  \textbf{86}(11),  2278--2324 (Nov 1998). \doi{10.1109/5.726791}, \url{https://ieeexplore.ieee.org/document/726791}, conference Name: Proceedings of the IEEE

\bibitem{cramer_heidelberg_2022}
Cramer, B., Stradmann, Y., Schemmel, J., Zenke, F.: The {Heidelberg} {Spiking} {Data} {Sets} for the {Systematic} {Evaluation} of {Spiking} {Neural} {Networks}. IEEE Transactions on Neural Networks and Learning Systems  \textbf{33}(7),  2744--2757 (Jul 2022). \doi{10.1109/TNNLS.2020.3044364}, \url{https://ieeexplore.ieee.org/document/9311226}, conference Name: IEEE Transactions on Neural Networks and Learning Systems

\bibitem{banner_scalable_2018}
Banner, R., Hubara, I., Hoffer, E., Soudry, D.: Scalable {Methods} for 8-bit {Training} of {Neural} {Networks} (Jun 2018). \doi{10.48550/arXiv.1805.11046}, \url{http://arxiv.org/abs/1805.11046}, arXiv:1805.11046

\bibitem{you_shiftaddnet_2020}
You, H., Chen, X., Zhang, Y., Li, C., Li, S., Liu, Z., Wang, Z., Lin, Y.: {ShiftAddNet}: {A} {Hardware}-{Inspired} {Deep} {Network} (Oct 2020). \doi{10.48550/arXiv.2010.12785}, \url{http://arxiv.org/abs/2010.12785}, arXiv:2010.12785

\bibitem{davies_loihi_2018}
Davies, M., Srinivasa, N., Lin, T.H., Chinya, G., Cao, Y., Choday, S.H., Dimou, G., Joshi, P., Imam, N., Jain, S., Liao, Y., Lin, C.K., Lines, A., Liu, R., Mathaikutty, D., McCoy, S., Paul, A., Tse, J., Venkataramanan, G., Weng, Y.H., Wild, A., Yang, Y., Wang, H.: Loihi: {A} {Neuromorphic} {Manycore} {Processor} with {On}-{Chip} {Learning}. IEEE Micro  \textbf{38}(1),  82--99 (Jan 2018). \doi{10.1109/MM.2018.112130359}, \url{https://ieeexplore.ieee.org/document/8259423/}

\bibitem{gonzalez2024spinnaker2}
Gonzalez, H.A., Huang, J., Kelber, F., Nazeer, K.K., Langer, T., Liu, C., Lohrmann, M., Rostami, A., Sch{\"o}ne, M., Vogginger, B., et~al.: Spinnaker2: A large-scale neuromorphic system for event-based and asynchronous machine learning. arXiv preprint arXiv:2401.04491  (2024)

\end{thebibliography}

\end{document}